\documentclass{article}

\usepackage{arxiv}

\usepackage[utf8]{inputenc} % allow utf-8 input
\usepackage[T1]{fontenc}    % use 8-bit T1 fonts
\usepackage{hyperref}       % hyperlinks
\usepackage{url}            % simple URL typesetting
\usepackage{booktabs}       % professional-quality tables
\usepackage{amsfonts}       % blackboard math symbols
\usepackage{nicefrac}       % compact symbols for 1/2, etc.
\usepackage{microtype}      % microtypography
\usepackage{lipsum}		% Can be removed after putting your text content
\usepackage{graphicx}
\usepackage{natbib}
\usepackage{doi}
\usepackage{amsmath}
\usepackage{multirow}
\DeclareMathOperator*{\argmax}{arg\,max}

\title{Mapping Transformer Leveraged Embeddings for Cross-Lingual Document Representation}

\author{ \href{https://orcid.org/0000-0002-4498-2486}{\includegraphics[scale=0.06]{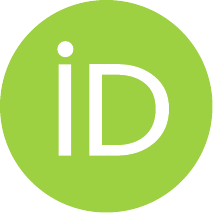}\hspace{1mm}Tsegaye Misikir Tashu} \\
	Department of Artificial Intelligence\\
	University of Groningen\\
	Groningen, 9747AG \\
	\texttt{t.m.tashu@rug.nl} \\
	\And
	\href{https://orcid.org/0009-0000-3291-3021}{\includegraphics[scale=0.06]{orcid.pdf}\hspace{1mm}Eduard-Raul Kontos} \thanks{The result presented in this work is based on Eduard-Raul Kontos's bachelor project while he was at the University of Groningen}\\
	Department of Artificial Intelligence\\
	University of Groningen\\
	Groningen, 9747AG \\
	\texttt{ediraul2001@gmail.com} \\
	    \AND
	 \href{https://orcid.org/0009-0007-7540-8616}{\includegraphics[scale=0.06]{orcid.pdf}\hspace{1mm} Matthia Sabatelli} \\
	Department of Artificial Intelligence\\ 
	University of Groningen\\
	Groningen, 9747AG \\
	\texttt{m.sabatelli@rug.nl} \\
	\And
	\href{https://orcid.org/0000-0001-5793-9498}{\includegraphics[scale=0.06]{orcid.pdf}\hspace{1mm}Matias Valdenegro-Toro} \\
	Department of Artificial Intelligence\\
	University of Groningen\\
	Groningen, 9747AG \\
	\texttt{m.a.valdenegro.toro@rug.nl} \\
}

\renewcommand{\shorttitle}{Transformer Leveraged Document Representation}

\hypersetup{
pdftitle={Mapping Transformer Leveraged Embeddings for Cross-Lingual Document Representation},
pdfsubject={q-bio.NC, q-bio.QM},
pdfauthor={David S.~Hippocampus, Elias D.~Striatum},
pdfkeywords={First keyword, Second keyword, More},
}

\begin{document}
\maketitle

\begin{abstract}
Recommendation systems, for documents, have become tools to find relevant content on the Web. However, these systems have limitations when it comes to recommending documents in languages different from the query language, which means they might overlook resources in non-native languages. This research focuses on representing documents across languages by using Transformer Leveraged Document Representations (TLDRs) that are mapped to a cross-lingual domain. Four multilingual pre-trained transformer models (mBERT, mT5 XLM RoBERTa, ErnieM) were evaluated using three mapping methods across 20 language pairs representing combinations of five selected languages of the European Union. Metrics like Mate Retrieval Rate and Reciprocal Rank were used to measure the effectiveness of mapped TLDRs compared to non-mapped ones. The results highlight the power of cross-lingual representations achieved through pre-trained transformers and mapping approaches suggesting a promising direction for expanding beyond language connections, between two specific languages.

\end{abstract}

% keywords can be removed

\section{Introduction}

The rapid expansion of online information from diverse sources and the growing multilingual nature of the web underscore the escalating significance of information retrieval (IR) and recommender systems (RS). Today's web is no longer limited to a single language, but is increasingly rich in multiple languages, mirroring the multilingual capacities of its global users \cite{Steichen2014,nccTashu}. This diversity highlights the urgent need for cross-lingual recommender systems. Traditional recommender systems often prioritize content in a single language, sidelining a wealth of multilingual documents that may hold valuable insights. This gap leads to the emergence of cross-language information access, where recommender systems suggest items in different languages based on user queries \cite{Lops2010,NARDUCCI2016,LCA}.

Machine Learning and Deep Learning, which have significantly impacted language representation and processing, are pivotal to enhancing information retrieval and recommender systems, especially in the realm of document recommendation \cite{nccTashu,feng-etal-2022}. With these advancements, documents ranging from historical texts and scientific papers to legal ones can be recommended more accurately. However, current recommender systems falter when content is available in various languages, often recommending documents in only the query language. In multinational contexts such as the European Union, such limitations can hinder effective policy formation.

There are two main strategies to address this gap: on the one hand, one can translate the query into multiple target languages or develop a cross-lingual representation space for documents. While this can be effective, this approach is fraught with challenges, including the need for large-scale data, the computational expense of training, and potential loss in translation, especially in domains like law that require precision. On the other hand, cross-lingual representations, which focus on creating shared embedding spaces for documents across languages, are the focal point of this study \cite{nccTashu}. By employing mapping-aligned document embeddings and comparing their similarity with the query, it offers a computationally cheaper solution without the need for extensive fine-tuning of pre-trained large language models.

The rest of the paper is organized as follows. Section \ref{sec:related} presents the related works. The proposed methodology is presented in section \ref{sec:proposed}. Section \ref{sec:experiment} presents the experimental setting and the datasets used in this work. The experimental results will be presented in Section \ref{sec:results}, while the results are discussed in Section \ref{sec:discussion}. Finally, the conclusions will be presented in section \ref{sec:conclusion}.

\section{related work}
\label{sec:related}

The work towards generating inter-lingual and multilingual representations, which can encapsulate information across multiple languages in a unified form, has gained substantial attention in recent years. This interest spans both word-level and document-level representations. Early observations, such as those introduced by \citep{mikolov2013exploiting}, identified that word embedding spaces across languages possess structural similarities. These insights led to the development of linear mappings from one language embedding space to another, utilizing parallel vocabularies. Subsequent works \cite{conneau2017word,smith2017offline,xing2015normalized}, have aimed to refine these cross-lingual word embeddings, mainly through modifications in space alignment methods or retrieval techniques.
Techniques like averaging word vectors \cite{litschko2018unsupervised} or leveraging cross-lingual knowledge bases like Wikipedia \cite{potthast2008wikipedia} or BabelNet \cite{franco2014knowledge} have been used to learn document-level cross-lingual representation. A notable methodology in this domain is the cross-lingual semantic indexing (CL-LSI) \cite{deerwester1990indexing, saad2014cross}, which extends the well-known latent semantic indexing (LSI) to encapsulate multiple languages through the singular value decomposition of concatenated monolingual document-term matrices.

An emerging strategy in both word-level, sentence-level and document-level research is the use of neural network architectures. One of the pioneer works in this direction was the work by \cite{schwenk2017learning} where they used a deep neural network to directly encode long text passages in a language-independent manner. The work by \cite{artetxe-schwenk-2019} used a multilingual auto-encoder to generate language-independent sentence embeddings. 
Recently, pre-trained models such as BERT \cite{devlin2018bert} have changed the landscape of cross-lingual representation research. These models have enabled the generation of sentence encoders on multilingual unlabeled corpora without the need for parallel data \cite{wu2019emerging,feng-etal-2022,goswami-etal-2021,Litschko}. Concurrently, certain studies have leveraged pre-trained multilingual transformers for cross-lingual information retrieval (IR). The work by \citep{XLingualmberttrans} combined mBERT with Google Translate in their information retrieval pipeline, while \citet{Litschko} utilized mBERT and XLM for the same purpose, emphasizing the need for fine-tuning for efficient and effective document-level results.  Collectively, these studies underscore the potential of transformers in cross-lingual information retrieval, paving the way for alternative methodologies such as mapping over fine-tuning, as explored in the current investigation.
While these approaches have shown promise, the study herein differentiates itself by presenting a methodology that uses mapping methods to create inter-lingual representations. The novelty of this work primarily lies in the use of mapping methods to align monolingual representations obtained separately for each language from pre-trained large language models, to produce inter-lingual document-level representations.

\section{Methods}
\label{sec:proposed}
In this section, we will introduce the different large language models used in this study and the mapping approaches used to learn interlingual representation from the pre-trained large language models.
%This project's methodology consisted of 4 main components: (1) the data, (2) transformer models, (3) embeddings, and (4) mapping approaches. Each aspect will be thoroughly discussed in the following subsections.

\subsection{Transformers}\label{sec:transformers}

\begin{figure}[h]
    \centering
    \includegraphics[width=0.4\textwidth]{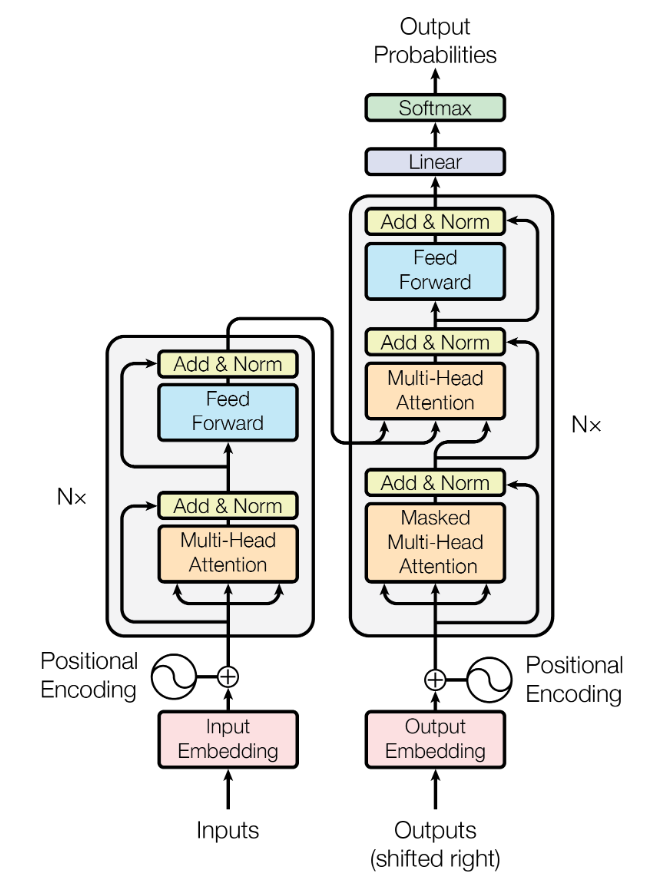}
    \caption{Diagram of the original transformer architecture. Image courtesy of \cite{Vaswani}.}
    \label{fig:trans}
\end{figure}

Transformers, shown in figure \ref{fig:trans} and introduced by Vaswani et al. (\citeyear{Vaswani}) have transformed the landscape of natural language processing (NLP). Instead of relying heavily on recurrent or convolutional layers, transformers leverage multiple attention heads to weigh the significance of different parts of an input sequence differently, allowing for parallel processing and the capture of long-range dependencies in data.  There exist a plethora of variations within the transformer architecture. In the following sections, we will discuss the specific variants of transformer-based large language models used in the context of this study.

\subsubsection{mBERT}

Multilingual BERT is an extension of the Bidirectional Encoder Representation from Transformers (BERT) that was introduced by Devlin et al. \cite{devlin2018bert}. BERT stands out as a pre-trained model, having undergone training on vast volumes of unlabelled data, primarily focusing on two pre-training objectives:

\begin{itemize}
    
\item \textbf{Masked Language Modelling (MLM):} This objective requires the model to predict masked portions of the provided input. Specifically, $15\%$ of the training data tokens undergo masking. Of these masked tokens, $80\%$ are substituted with the "[MASK]" placeholder, $10\%$ are replaced with a random token, and the remaining $10\%$ are left unaltered.

\item \textbf{Next Sentence Prediction (NSP):} BERT's versatility allows it to manage tasks that involve pairs of sentences, which may or may not exhibit contextual coherence. During its training phase, BERT was supplied with sentence pairs where $50\%$ of the pairs were contextually sequential from the training dataset, while the remaining $50\%$ constituted random, unrelated sentences.

\end{itemize} 

BERT was originally pre-trained on a strictly monolingual English corpus. Recognizing the limitations of such a unilingual approach, there emerged a demand for a model with broader linguistic capabilities. In response, the Multilingual BERT (mBERT) \cite{devlin2018bert}, was conceptualized. This iteration extends the foundational principles of BERT, accommodating text from a diverse array of 104 languages.

\subsubsection{mT5}

Multilingual\citep{mT5} Text-to-Text Transfer Transformer (mT5) is an encoder-decoder model pre-trained on 101 languages, closely based on the original T5 model from \citep{T5}. It has been pre-trained on an objective similar to MLM, called MLM span-corruption, where consecutive tokens from the input are masked from the model during pre-training. %mT5 is similar in architecture to BERT (also closely resembling the original transformer architecture), with the difference being that mT5 uses one training objective compared to BERT's 2. Furthermore, it is an encoder-decoder model, while BERT is only an encoder model.

mT5 is highly specialised for text-to-text tasks such as machine translation and text generation, however, it can also be used as an encoder model only, which was done for this project. Like BERT, the maximum amount of tokens that were used was $512$, with an embedding dimensionality of $768$, corresponding to the "base" version.

\subsubsection{XLM-RoBERTa}

The Cross-Lingual Modelling for Robustly Optimised BERT, colloquially termed XLM-RoBERTa, stands as a notable iteration of pre-trained multilingual transformers. Introduced by \citeauthor{XLMR} \citeyearpar{XLMR}, this model is an evolution of RoBERTa \citep{roberta}. Diverging from conventional methodologies, XLM-RoBERTa eschews both the Next Sentence Prediction (NSP) and translation objectives, concentrating exclusively on Masked Language Modelling (MLM). The key innovation lies in refining the training procedure and extending the training duration, measures that synergistically enhance model performance. Adapted to cater to 100 languages, XLM-RoBERTa can function effectively as an encoder-only model. For the purposes of this research, the "base" variant of XLM-RoBERTa was deployed, accommodating a maximum of $512$ tokens and featuring an embedding dimensionality of $768$.

\subsubsection{ErnieM}

The Multilingual Ernie (ErnieM) \citep{erniem} represents a distinguished pre-trained multilingual transformer. Drawing inspiration from the XLM-RoBERTa, ErnieM's hallmark feature lies in its capacity to synchronize linguistic representations across its embedded languages. This harmonization is operationalized through a cross-lingual semantic alignment, juxtaposing parallel data with its monolingual counterpart. In the spirit of achieving this, the authors put forth two pre-training objectives:

\begin{itemize}
\item Cross-Attention MLM (CAMLM): A strategy devised to cohesively align the semantic representation of parallel data across the entire linguistic spectrum.
\item Back-Translation MLM (BTMLM): This objective embarks on aligning cross-lingual semantics with monolingual contexts. Through back-translation, it facilitates the generation of novel linguistic tokens from monolingual corpora, and subsequently acquaints the model with their multilingual semantic alignment.
\end{itemize}

Supplemented by the translation modelling language task (an initiative akin to MLM but marked by the amalgamation of sequences from an array of languages) and the Multilingual MLM (characterized by masking tokens transcending diverse languages), these objectives jointly constitute the pre-training paradigm of ErnieM. Maintaining consistency, this study harnesses the "base" version of ErnieM, with a stipulated threshold of $512$ tokens and an embedding dimensionality set at $768$.

The models selected for this investigation inherently embrace a multilingual ethos, underpinned by two pivotal reasons: Firstly, the monolingual iterations of these models have not ubiquitously undergone training across the selected quintet of languages earmarked for this research. More critically, the inherent overlap in the models' embedding space across languages posits a fertile ground to evaluate the potential of leveraging ready-made multilingual models sans the requisite of supplementary mapping or precision-tuning. To draw an illustrative parallel, juxtaposing disparate models of analogous frameworks, each tailored to individual languages (e.g., BERT vis-à-vis its Gallic analogue), might yield embeddings that, owing to divergent training trajectories, manifest disparities too profound to be semantically reconciled.

% As mentioned in the previous sections, only the encoder part of the models was used, as this project only aims to use the embeddings generated by the last hidden state and not generate any additional text from the models. Furthermore, all of them were used in the same manner: a tokeniser and model instance were initialised from the model's weights.

\subsection{Mapping approaches}\label{sec:mapping}
Given two monolingual document collections, $D_{x}=\{d_{x,1}, \dots, d_{x,n}\}$ in language $x$ and $D_{y}=\{d_{y,1}, \dots, d_{y,n}\}$ in language $y$ . To embark on a nuanced analysis of these documents, it is imperative first to learn or extract the embedding for each document. To achieve this, we employ the pre-trained large language models introduced in section $3$ subsection $3.1$.  Notwithstanding, it's worth noting that any representation learning algorithm that embeds the document sets $D_{x}$ and $D_{y}$ into vectors within the space $\mathbb{R}^{k}$ can be used.

From the language models, we obtain sets of vectors, respectively, defined as $C_{x}=\{\hat{d}_{x,1}, \dots, \hat{d}_{x,n}\} \subset \mathbb{R}^{k}$ and $C_{y}=\{\hat{d}_{y,1}, \dots, \hat{d}_{y,n}\} \subset \mathbb{R}^{k}$.  Conceptually, $C_{x}$ and $C_{y}$ can be interpreted as "Conceptual Vector Spaces", encapsulating broader linguistic and thematic abstractions inherent to the original documents. Nevertheless, a salient point to recognize is that even if vectors within $C_{x}$ and $C_{y}$ encapsulate analogous concepts transversal to languages, the representation schema might vary. Consequently, a mere direct juxtaposition of $\hat{d}_{x,k}$ and $\hat{d}_{y,k}$might not manifest the underlying content congruencies.

All the mapping methods used in this study are adopted from the works of Tashu et al.\citep{nccTashu}.  In the upcoming section, we will present a summary of three different mappings where more details on each of the methods can be found in \citep{Marc2021,nccTashu}.

\subsubsection{Linear concept approximation (LCA)}

The motivation is to directly embed the test documents into the space spanned by the training documents in the semantic space using linear least squares \cite{LCA}. This is based on the assumption that the vector space spanned by the parallel training documents is the same in their respective language. Therefore, the coordinates of the test documents in that span would be a good language-independent representation of these documents. Using the representation obtained from the large language models presented in section 3, we can derive low-dimensional representations of each document within $\mathbb{R}^{k}$. Multiple documents can be concatenated into matrices.  If there are $n$ documents available in both languages, we can create the representation/concept matrices ${C}_x = {X}^T \in \mathbb{R}^{n \times k}$ and ${C}_y = {Y}^T \in \mathbb{R}^{n \times k}$ in which every column is a concept in its respective language.

\subsubsection{Linear Concept Compression (LCC)}

The motivation behind LCC is to find mappings into an inter-lingual space, $E
C_{x}, C_{y}$, such that the comparison of $C_{x}(\hat{d}_{x,k}), C_{y}(\hat{d}_{y,k}),$ provides a measure of content similarity. For two monolingual representations, we want to find their inter-lingual representations, which encode the same information as the different monolingual spaces do. More precisely, for a given document $d$ and its representations in each respective language, $\hat{d}_{x,k}$ and $\hat{d}_{y,k}$, we want to find mappings $C_{x}$ and $C_{y}$, respectively, such that $C_{x}(\hat{d}_{x,k}) = C_{y}(\hat{d}_{y,k})$ and the information of $\hat{d}_{x,k}$ and $\hat{d}_{y,k}$ is preserved.  The intuition is to train an Encoder-Decoder approach. The purpose of the Encoder is to encode monolingual representations in a language-independent space. The purpose of the Decoder is to reconstruct the monolingual representations of multiple languages from that encoding\cite{Map}. 

\subsubsection{Neural Concept Approximation (NCA)}

In contrast to conventional approaches where mappings are directly derived from given vectors, $C_x$ and $C_y$, the proposed methodology leverages a Neural Network to approximate these vectors. Specifically, a Feed Forward Neural Network (FFNN). Two distinct models were trained: one mapping from the source language to the target language, and the other in the reverse direction \cite{nccTashu}.

% Unlike the other approaches, where the mapping is built directly from some given ${C_x}$ and ${C_y}$, this method attempts to approximate them using a Neural Network. This has been done using a Feed Forward Neural Network (FFNN) and training two separate models: 1 from the source language to the target language and vice-versa.

Both models were defined in the same manner: 1 layer of $500$ neurons, using the Exponential Linear Unit (ELU), with the Huber objective function, for a maximum of $250$ epochs with the implementation of early stopping and a learning rate of $5\cdot 10^{-4}$. The network's architecture consists of 3 total layers, one input layer with dimensionality $d$ (the dimension of a given document), followed by the hidden layer (with dimensionality $d\times 500$), and the output layer with dimensionality $500\times d$.

%The training data consists of $75\%$ of the initial docments, and $25\%$ of the data is used for testing. After training, the models are used to predict the test data, more specifically predicting what the output vector of a given test document should be. This is concatenated with the testing data of the target language (or source language) and given to an evaluation metric. After creating the general projections, it was necessary to test their performance. This was achieved by mapping a set of unseen test data, either from S if the mapping was done from the source to the target or from T if it was from the target to the source. To quantify the performance of the models, it was necessary to use two evaluation metrics.

\section{Experiment}
\label{sec:experiment}

\subsection{Data}\label{sec:data}

The JRC-Acquis corpus \citep{JRC-Acquis} was used for this project because of its characteristics. It is a publicly available, sentence-aligned corpus consisting of the 22 official languages of the European Union (EU), containing legal documents pertaining to EU matters from 1958 to 2006. 
Since this study dealt with language pairs, only five languages were used, those being English, Romanian, Dutch, German, and French, for a total of 20 ordered pairs (i.e. English $\rightarrow$ French and French $\rightarrow$ English are treated as a different pair). Since the documents for each language were not aligned, it was necessary to perform a secondary alignment for the five chosen languages such that documents were shared across the subset, resulting in $6,538$ unique documents. There were also some issues at the character level of some non-English documents from the initial dataset. For instseveralber of French documents presented corrupted letters, meaning that letters with diacritics were instead displayed in XML format (e.g. "é" displayed as "\%eacute"). A preprocessing step was as such introduced to replace these corrupted variants with their original form and to remove any additional white space from the documents. The documents, at the same time, were converted from XML to a standard string format to be used by the models.  In this study, 60\% was used for the training set, 20\% for the validation set and 20\% for the test set. 

\subsection{Embeddings}\label{sec:emb}

It is necessary to represent the documents in a continuous manner to be able to apply any mapping approach. This was achieved by passing all documents, in each language, through the tokenizer and model modules of the previously discussed transformer models.

An input text undergoes several processing steps while passing through the tokenizer: it is truncated or padded to the maximum length allowed by the models ($N=512$ tokens), after which the tokens are converted to internal ID representations stored in the vocabulary of the model, and for which the attention mask is computed. The latter part allows the model to look only at the relevant tokens in the sequence, ignoring padding tokens. Since this study only deals with the embeddings of the models and not their decoded outputs, the final hidden state from the encoder part of the models is extracted. The model computed the embedding for each token, and as such, documents are now represented as $512\times 768$ matrices, while it is necessary to obtain a vector of size $768$. This was solved by performing a global pooling operation on all of the outputted states, where tokens that were not ignored by the attention mask were averaged together. As such, documents are now represented by vectors with dimensionality $768$, to be used in the following section.

\subsection{Evaluation metrics}

Two evaluation metrics were used to compute the performance of the mapping approaches:

\begin{itemize}
    \item Mate Retrieval Rate: the retrieval rate of the most symmetric document; this metric evaluates how similar two documents are - the query and retrieved document. If the retrieved document is the same as the query document, that is called a mate retrieval. It is defined as:
    \begin{equation}\label{eq:auxMate}
        \begin{aligned}
            MR(d) &= \underset{}{\argmax}\ \textbf{S}_d\cdot \textbf{T}_d^T\\
            S(d, d') &= \begin{cases}
                1 & d=d' \\
                0 & d\neq d'
                \end{cases}
        \end{aligned}
    \end{equation}

    where $S$ is the similarity between 2 documents $d$ and $d'$, and $MR$ is the mate retrieval for a given document $d$ in the source $S$ and target language $T$. It can be said that a mate retrieval is successful if $d$ and $d'$ are the same. The equations in \ref{eq:auxMate} can be combined to compute the mate retrieval rate for all documents ($D$), as seen in equation \ref{eq:retrieval}:

    \begin{equation}\label{eq:retrieval}
        \text{RetrievalRate} = \frac{1}{|D|}\sum_{d=1}^{|D|} S(d, MR(d))
    \end{equation}
    
    \item Mean Reciprocal Rank: this represents how high-ranked documents are, based on a similarity measure. This has been achieved using cosine similarity, defined below:
    \begin{equation}
        C(d_1, d_2) = \frac{{d}_1\cdot{d}_2}{\|{d}_1\|\cdot \| {d}_2\|}
    \end{equation}

    where the numerator represents the inner product of the vector representations of documents $d_1$ and $d_2$, and the denominator is the magnitude product of the two vectors. If the two documents are similar to each other, their cosine will be closer to $1$ and will be closer to $-1$ if they are not similar. This equation can be used to obtain the cosine matrix similarity of all documents.
    
    Furthermore, the rank $r$ of a document can be defined as its cosine similarity compared to other documents obtained from the matrix cosine similarity. If it is most similar to itself in the target language, then its rank will be $1$. Finally, these components can be combined to form the mean reciprocal rank:
    \begin{equation}
        \text{ReciprocalRank}=\frac{1}{|D|}\sum_{d=1}^{|D|}\frac{1}{r_d}\
    \end{equation}
    
\end{itemize}

\section{Results}
\label{sec:results}

\begin{figure*}
    \centering
    \includegraphics[width=\textwidth]{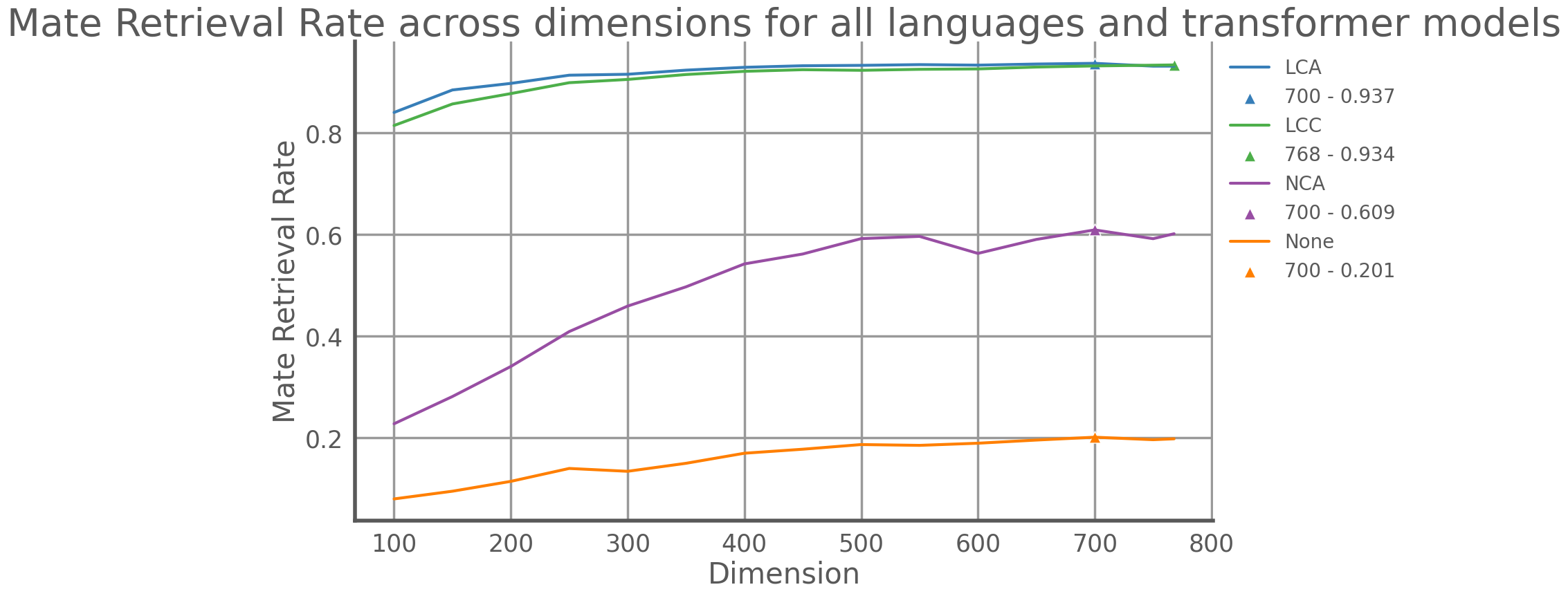}
    \caption{Lineplot of the average Mate Retrieval Rate across dimensions for all language pairs and models, using LCA, LCC, NCA, and no mapping.}
    \label{fig:rrate}
\end{figure*}

\begin{figure*}
    \centering
    \includegraphics[width=\textwidth]{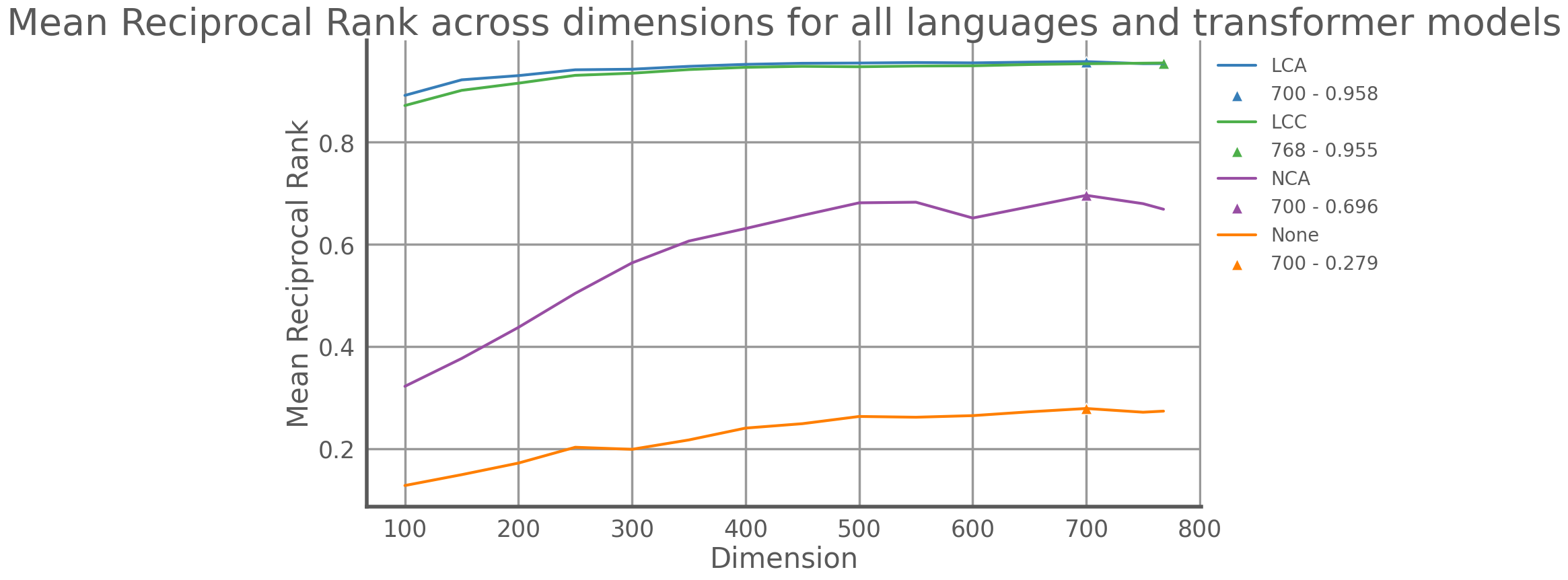}
    \caption{Lineplot of the average Mean Reciprocal Rank across dimensions for all language pairs and models, using LCA, LCC, NCA, and no mapping.}
    \label{fig:rrank}
\end{figure*}

The performance of the mapped (or not) embeddings was measured using the evaluation metrics defined in the previous section. Due to the large number of results that were obtained ($640$ total results across four transformer models, three mapping methods and no mapping, for 20 language pairs, for each evaluation metric), the final results have been averaged across models and language pairs. As such, Figures \ref{fig:rrate} and \ref{fig:rrank} only present their average evaluation metric for all dimensions. Both figures showcase significant results when comparing mapped and non-mapped embeddings. However, there is also a significant difference between embeddings mapped using NCA and embeddings mapped with the other methods. 

\begin{table*}[h!]
    \centering
    \begin{tabular}{c|c|c|c}
        Model & Mapping & Mean Reciprocal Rank & Mate Retrieval Rate\\ \hline
        \multirow{4}{*}{mBERT} & None & 0.2 & 0.115\\
        & LCA* & \textbf{0.975} & \textbf{0.963} \\
        & LCC & 0.973 & 0.959 \\
        & NCA & 0.84 & 0.781 \\ \hline
        \multirow{4}{*}{mT5} & None & 0.466 & 0.37 \\
        & LCA* & 0.947 & 0.922 \\
        & LCC & 0.936 & 0.907 \\
        & NCA & 0.814 & 0.756 \\ \hline
        \multirow{4}{*}{XLM-RoBERTa} & None & 0.114 & 0.057 \\
        & LCA & 0.948 & 0.925 \\
        & LCC* & 0.951 & 0.928 \\
        & NCA & 0.617 & 0.499 \\ \hline
        \multirow{4}{*}{ErnieM} & None & 0.443 & 0.355 \\
        & LCA* & 0.965 &  0.949 \\
        & LCC & 0.962 & 0.946 \\
        & NCA & 0.742 & 0.67 \\
      \hline
    \end{tabular}
    \caption{Mean Reciprocal Rank and Mate Retrieval Rate.}
    \label{tab:results_mean}
\end{table*}

The best mapping method across both evaluation metrics was LCA ($\text{Retrieval Rate}=0.937$, $\text{Reciprocal Rank}=0.958$), while the worst mapping method was NCA ($\text{Retrieval Rate}=0.609$, $\text{Reciprocal Rank}=0.696$). Still, all methods performed significantly better than the non-mapped embeddings ($\text{Retrieval Rate}=0.201$, $\text{Reciprocal Rank}=0.279$).

Table \ref{tab:results_mean} presents the results across all language pairs for both metrics, broken down for each transformer model and mapping method, and additionally the results obtained by \citeauthor{nccTashu} \citeyearpar{nccTashu}. Using the same mapping approaches, mBERT embeddings mapped using LCA outperform all other models and mapping combinations, including those from the mentioned paper, across both metrics ($\text{RetrievalRate}=0.963$, $\text{ReciprocalRank}=0.975$).

\section{Discussion
}\label{sec:discussion}

From our results, it becomes evident that Transformer Leveraged embeddings combined with mapping methods markedly outperform non-mapped embeddings across all models, as delineated in Table \ref{tab:results_mean}. These Leveraged embeddings, in all instances, show significant superiority compared to the non-mapped variants. This underscores that employing an off-the-shelf model devoid of enhancements (e.g., fine-tuning, mapping) results in subpar outcomes, irrespective of the model's type. Figures \ref{fig:rrate} and \ref{fig:rrank} further substantiate this, demonstrating that mapped embeddings consistently outpace their non-mapped counterparts across all metrics. Within this context, the NCA mapping method displayed the favourableorable performance, overshadowing only the non-mapped embeddings. This could be attributable to the network's architectural design, potentially falling short in capturing the nuanced similarities between documents to establish an effective mapping.

An examination of Table \ref{tab:results_mean} reveals mBERT's dominance over other transformer models across all mapping strategies. Notably, when paired with LCA and LCC-mapped embeddings, mBERT eclipsed all other embedding and mapping combinations as referenced by \citep{nccTashu}. This superior performance may be credited to the extensive data mBERT trained on, complemented by its pre-training tasks. 

Interestingly, both ErnieM and mT5, when aligned with non-mapped embeddings, showcased better performance than their transformer counterparts under identical conditions. The underlying reason might be traced back to the distinctive training data and methodologies employed by these models. In contrast to mBERT and XLM-RoBERT, which utilize MLM (and additionally NSP for mBERT), ErnieM incorporates a broader spectrum of pre-training objectives geared towards cross-lingual alignment. This distinction could elucidate the superior performance of its non-mapped embeddings. mT5's commendable performance can be attributed to its foundational design, being inherently an encoder-decoder model, though this project exclusively utilized its encoding facet.

In general, our study highlights the efficacy of Transformer Leveraged embeddings when synergized with mapping techniques, resulting in a noticeable performance leap over non-mapped embeddings. This aligns with the findings of \citep{Litschko}, which accentuate that standalone out-of-the-box models, without refinements or supplementary techniques, are generally less efficient. However, diverging from their study, our research underscores that optimal performance doesn't solely hinge on model fine-tuning. In the realm of IR, integrating mapping techniques can be equally potent in driving commendable results.

\section{Conclusion}
\label{sec:conclusion}

Document recommendation stands at the forefront of Information Retrieval (IR) systems. Within recommendation frameworks, it efficiently suggests pertinent documents in alignment with a user's query. In our research, we delved into the possibilities of crafting cross-lingual representations by harnessing embeddings from pre-existing multilingual transformers in conjunction with mapping strategies. Using embeddings from these pre-trained multilingual transformers allows for document representation without requiring further training or intricate processing. Nonetheless, our research illuminated that solely depending on the raw embeddings from the transformers fell short in terms of efficacy. A notable enhancement in results was witnessed when the embeddings were synergized with mapping techniques such as LCA, LCC, and NCA.
The languages incorporated within our study hold considerable prominence across various linguistic tasks. Consequently, the adopted models and mapping techniques have the potential to foster efficient representations by mapping low-resource languages onto those that are more abundantly represented. It beckons further exploration into how these mapping techniques perform when applied to low-resource languages. Future research might not restrict itself to merely language pairs, as was the focus of this study, but could expand to encompass language tuples—translating from a single source language to multiple target languages. Achieving this might necessitate refining the present mapping methodologies, introducing supplementary steps, or pioneering entirely novel methods. The code of this project is publicly available on \href{https://github.com/Tron404/BScThesis.git}{GitHub}.

\bibliographystyle{unsrtnat}
\bibliography{references}  %%% Uncomment this line and comment out the ``thebibliography'' section below to use the external .bib file (using bibtex) .

\begin{thebibliography}{31}
\providecommand{\natexlab}[1]{#1}
\providecommand{\url}[1]{\texttt{#1}}
\expandafter\ifx\csname urlstyle\endcsname\relax
  \providecommand{\doi}[1]{doi: #1}\else
  \providecommand{\doi}{doi: \begingroup \urlstyle{rm}\Url}\fi

\bibitem[Steichen et~al.(2014)Steichen, Ghorab, O'Connor, Lawless, and Wade]{Steichen2014}
Ben Steichen, M.~Rami Ghorab, Alexander O'Connor, S{\'e}amus Lawless, and Vincent Wade.
\newblock Towards personalized multilingual information access - exploring the browsing and search behavior of multilingual users.
\newblock In Vania Dimitrova, Tsvi Kuflik, David Chin, Francesco Ricci, Peter Dolog, and Geert-Jan Houben, editors, \emph{User Modeling, Adaptation, and Personalization}, pages 435--446, Cham, 2014. Springer International Publishing.
\newblock ISBN 978-3-319-08786-3.

\bibitem[Tashu et~al.(2023)Tashu, Lenz, and Horv{\'a}th]{nccTashu}
Tsegaye~Misikir Tashu, Marc Lenz, and Tom{\'a}{\v s} Horv{\'a}th.
\newblock {{NCC}}: {{Neural}} concept compression for multilingual document recommendation.
\newblock \emph{Applied Soft Computing}, 142:\penalty0 110348, July 2023.
\newblock ISSN 15684946.
\newblock \doi{10.1016/j.asoc.2023.110348}.

\bibitem[Lops et~al.(2010)Lops, Musto, Narducci, De~Gemmis, Basile, and Semeraro]{Lops2010}
Pasquale Lops, Cataldo Musto, Fedelucio Narducci, Marco De~Gemmis, Pierpaolo Basile, and Giovanni Semeraro.
\newblock Mars: A {M}ultil{A}nguage {R}ecommender {S}ystem.
\newblock In \emph{Proceedings of the 1st International Workshop on Information Heterogeneity and Fusion in Recommender Systems}, HetRec '10, page 24–31, New York, NY, USA, 2010. ACM.
\newblock ISBN 9781450304078.

\bibitem[Narducci et~al.(2016)Narducci, Basile, Musto, Lops, Caputo, {de Gemmis}, Iaquinta, and Semeraro]{NARDUCCI2016}
Fedelucio Narducci, Pierpaolo Basile, Cataldo Musto, Pasquale Lops, Annalina Caputo, Marco {de Gemmis}, Leo Iaquinta, and Giovanni Semeraro.
\newblock Concept-based item representations for a cross-lingual content-based recommendation process.
\newblock \emph{Information Sciences}, 374:\penalty0 15--31, 2016.
\newblock ISSN 0020-0255.

\bibitem[Salamon et~al.(2021)Salamon, Tashu, and Horv\'{a}th]{LCA}
Vilmos~Tibor Salamon, Tsegaye~Misikir Tashu, and Tom\'{a}\v{s} Horv\'{a}th.
\newblock Linear concept approximation for multilingual document recommendation.
\newblock In \emph{Intelligent Data Engineering and Automated Learning – IDEAL 2021: 22nd International Conference, IDEAL 2021, Manchester, UK, November 25–27, 2021, Proceedings}, page 147–156, Berlin, Heidelberg, 2021. Springer-Verlag.
\newblock ISBN 978-3-030-91607-7.
\newblock \doi{10.1007/978-3-030-91608-4_15}.
\newblock URL \url{https://doi.org/10.1007/978-3-030-91608-4_15}.

\bibitem[Feng et~al.(2022)Feng, Yang, Cer, Arivazhagan, and Wang]{feng-etal-2022}
Fangxiaoyu Feng, Yinfei Yang, Daniel Cer, Naveen Arivazhagan, and Wei Wang.
\newblock Language-agnostic {BERT} sentence embedding.
\newblock In \emph{Proceedings of the 60th Annual Meeting of the Association for Computational Linguistics (Volume 1: Long Papers)}, pages 878--891, Dublin, Ireland, May 2022. Association for Computational Linguistics.
\newblock \doi{10.18653/v1/2022.acl-long.62}.
\newblock URL \url{https://aclanthology.org/2022.acl-long.62}.

\bibitem[Mikolov et~al.(2013)Mikolov, Le, and Sutskever]{mikolov2013exploiting}
Tom\'{a}s Mikolov, Quoc~V. Le, and Ilya Sutskever.
\newblock Exploiting similarities among languages for machine translation.
\newblock \emph{CoRR}, abs/1309.4168, 2013.
\newblock URL \url{http://arxiv.org/abs/1309.4168}.

\bibitem[Lample et~al.(2018)Lample, Conneau, Ranzato, Denoyer, and Jégou]{conneau2017word}
Guillaume Lample, Alexis Conneau, Marc'Aurelio Ranzato, Ludovic Denoyer, and Hervé Jégou.
\newblock Word translation without parallel data.
\newblock In \emph{International Conference on Learning Representations}, pages 1--14, 2018.
\newblock URL \url{https://openreview.net/forum?id=H196sainb}.

\bibitem[Smith et~al.(2017)Smith, Turban, Hamblin, and Hammerla]{smith2017offline}
Samuel~L Smith, David~HP Turban, Steven Hamblin, and Nils~Y Hammerla.
\newblock Offline bilingual word vectors, orthogonal transformations and the inverted softmax.
\newblock \emph{arXiv:1702.03859}, 2017.

\bibitem[Xing et~al.(2015)Xing, Wang, Liu, and Lin]{xing2015normalized}
Chao Xing, Dong Wang, Chao Liu, and Yiye Lin.
\newblock Normalized word embedding and orthogonal transform for bilingual word translation.
\newblock In \emph{Proceedings of the 2015 Conference of the North {A}merican Chapter of the Association for Computational Linguistics: Human Language Technologies}, pages 1006--1011, Denver, Colorado, May{--}June 2015. Association for Computational Linguistics.
\newblock \doi{10.3115/v1/N15-1104}.
\newblock URL \url{https://aclanthology.org/N15-1104}.

\bibitem[Litschko et~al.(2018)Litschko, Glava\v{s}, Ponzetto, and Vuli\'{c}]{litschko2018unsupervised}
Robert Litschko, Goran Glava\v{s}, Simone~Paolo Ponzetto, and Ivan Vuli\'{c}.
\newblock Unsupervised cross-lingual information retrieval using monolingual data only.
\newblock In \emph{The 41st International ACM SIGIR Conference on Research \& Development in Information Retrieval}, SIGIR '18, page 1253–1256, New York, NY, USA, 2018. Association for Computing Machinery.
\newblock ISBN 9781450356572.
\newblock \doi{10.1145/3209978.3210157}.
\newblock URL \url{https://doi.org/10.1145/3209978.3210157}.

\bibitem[Potthast et~al.(2008)Potthast, Stein, and Anderka]{potthast2008wikipedia}
Martin Potthast, Benno Stein, and Maik Anderka.
\newblock A wikipedia-based multilingual retrieval model.
\newblock In Craig Macdonald, Iadh Ounis, Vassilis Plachouras, Ian Ruthven, and Ryen~W. White, editors, \emph{Advances in Information Retrieval}, pages 522--530, Berlin, Heidelberg, 2008. Springer Berlin Heidelberg.
\newblock ISBN 978-3-540-78646-7.

\bibitem[Franco-Salvador et~al.(2014)Franco-Salvador, Rosso, and Navigli]{franco2014knowledge}
Marc Franco-Salvador, Paolo Rosso, and Roberto Navigli.
\newblock A knowledge-based representation for cross-language document retrieval and categorization.
\newblock In \emph{Proceedings of the 14th Conference of the {E}uropean Chapter of the Association for Computational Linguistics}, pages 414--423, Gothenburg, Sweden, April 2014. Association for Computational Linguistics.
\newblock \doi{10.3115/v1/E14-1044}.
\newblock URL \url{https://aclanthology.org/E14-1044}.

\bibitem[Deerwester et~al.(1990)Deerwester, Dumais, Furnas, Landauer, and Harshman]{deerwester1990indexing}
Scott Deerwester, Susan~T Dumais, George~W Furnas, Thomas~K Landauer, and Richard Harshman.
\newblock Indexing by latent semantic analysis.
\newblock \emph{Journal of the American society for information science}, 41\penalty0 (6):\penalty0 391--407, 1990.

\bibitem[Saad et~al.(2014)Saad, Langlois, and Sma{\"i}li]{saad2014cross}
Motaz Saad, David Langlois, and Kamel Sma{\"i}li.
\newblock Cross-lingual semantic similarity measure for comparable articles.
\newblock In \emph{International Conference on Natural Language Processing}, pages 105--115. Springer, 2014.

\bibitem[Schwenk and Douze(2017)]{schwenk2017learning}
Holger Schwenk and Matthijs Douze.
\newblock Learning joint multilingual sentence representations with neural machine translation.
\newblock In \emph{Proceedings of the 2nd Workshop on Representation Learning for {NLP}}, pages 157--167, Vancouver, Canada, August 2017. Association for Computational Linguistics.
\newblock \doi{10.18653/v1/W17-2619}.
\newblock URL \url{https://aclanthology.org/W17-2619}.

\bibitem[Artetxe and Schwenk(2019)]{artetxe-schwenk-2019}
Mikel Artetxe and Holger Schwenk.
\newblock Massively multilingual sentence embeddings for zero-shot cross-lingual transfer and beyond.
\newblock \emph{Transactions of the Association for Computational Linguistics}, 7:\penalty0 597--610, 2019.
\newblock \doi{10.1162/tacl_a_00288}.
\newblock URL \url{https://aclanthology.org/Q19-1038}.

\bibitem[Devlin et~al.(2019)Devlin, Chang, Lee, and Toutanova]{devlin2018bert}
Jacob Devlin, Ming-Wei Chang, Kenton Lee, and Kristina Toutanova.
\newblock {BERT}: Pre-training of deep bidirectional transformers for language understanding.
\newblock In \emph{Proceedings of the 2019 Conference of the North {A}merican Chapter of the Association for Computational Linguistics: Human Language Technologies, Volume 1 (Long and Short Papers)}, pages 4171--4186, Minneapolis, Minnesota, June 2019. Association for Computational Linguistics.
\newblock \doi{10.18653/v1/N19-1423}.

\bibitem[Conneau et~al.(2020)Conneau, Wu, Li, Zettlemoyer, and Stoyanov]{wu2019emerging}
Alexis Conneau, Shijie Wu, Haoran Li, Luke Zettlemoyer, and Veselin Stoyanov.
\newblock Emerging cross-lingual structure in pretrained language models.
\newblock In \emph{Proceedings of the 58th Annual Meeting of the Association for Computational Linguistics}, pages 6022--6034, Online, July 2020. Association for Computational Linguistics.
\newblock \doi{10.18653/v1/2020.acl-main.536}.

\bibitem[Goswami et~al.(2021)Goswami, Dutta, Assem, Fransen, and McCrae]{goswami-etal-2021}
Koustava Goswami, Sourav Dutta, Haytham Assem, Theodorus Fransen, and John~P. McCrae.
\newblock Cross-lingual sentence embedding using multi-task learning.
\newblock In \emph{Proceedings of the 2021 Conference on Empirical Methods in Natural Language Processing}, pages 9099--9113, Online and Punta Cana, Dominican Republic, November 2021. Association for Computational Linguistics.
\newblock \doi{10.18653/v1/2021.emnlp-main.716}.
\newblock URL \url{https://aclanthology.org/2021.emnlp-main.716}.

\bibitem[Litschko et~al.(2022)Litschko, Vuli\'{c}, Ponzetto, and Glava\v{s}]{Litschko}
Robert Litschko, Ivan Vuli\'{c}, Simone~Paolo Ponzetto, and Goran Glava\v{s}.
\newblock On cross-lingual retrieval with multilingual text encoders.
\newblock \emph{Inf. Retr.}, 25\penalty0 (2):\penalty0 149–183, jun 2022.
\newblock ISSN 1386-4564.
\newblock \doi{10.1007/s10791-022-09406-x}.
\newblock URL \url{https://doi.org/10.1007/s10791-022-09406-x}.

\bibitem[Shi et~al.(2020)Shi, Bai, and Lin]{XLingualmberttrans}
Peng Shi, He~Bai, and Jimmy Lin.
\newblock Cross-{{Lingual Training}} of {{Neural Models}} for {{Document Ranking}}.
\newblock In \emph{Findings of the {{Association}} for {{Computational Linguistics}}: {{EMNLP}} 2020}, pages 2768--2773, {Online}, November 2020. {Association for Computational Linguistics}.
\newblock \doi{10.18653/v1/2020.findings-emnlp.249}.

\bibitem[Vaswani et~al.(2017)Vaswani, Shazeer, Parmar, Uszkoreit, Jones, Gomez, Kaiser, and Polosukhin]{Vaswani}
Ashish Vaswani, Noam Shazeer, Niki Parmar, Jakob Uszkoreit, Llion Jones, Aidan~N. Gomez, Lukasz Kaiser, and Illia Polosukhin.
\newblock Attention is all you need.
\newblock \emph{CoRR}, abs/1706.03762, 2017.
\newblock URL \url{http://arxiv.org/abs/1706.03762}.

\bibitem[Xue et~al.(2020)Xue, Constant, Roberts, Kale, Al{-}Rfou, Siddhant, Barua, and Raffel]{mT5}
Linting Xue, Noah Constant, Adam Roberts, Mihir Kale, Rami Al{-}Rfou, Aditya Siddhant, Aditya Barua, and Colin Raffel.
\newblock mt5: {A} massively multilingual pre-trained text-to-text transformer.
\newblock \emph{CoRR}, abs/2010.11934, 2020.
\newblock URL \url{https://arxiv.org/abs/2010.11934}.

\bibitem[Raffel et~al.(2019)Raffel, Shazeer, Roberts, Lee, Narang, Matena, Zhou, Li, and Liu]{T5}
Colin Raffel, Noam Shazeer, Adam Roberts, Katherine Lee, Sharan Narang, Michael Matena, Yanqi Zhou, Wei Li, and Peter~J. Liu.
\newblock Exploring the limits of transfer learning with a unified text-to-text transformer.
\newblock \emph{CoRR}, abs/1910.10683, 2019.
\newblock URL \url{http://arxiv.org/abs/1910.10683}.

\bibitem[Conneau et~al.(2019)Conneau, Khandelwal, Goyal, Chaudhary, Wenzek, Guzm{\'{a}}n, Grave, Ott, Zettlemoyer, and Stoyanov]{XLMR}
Alexis Conneau, Kartikay Khandelwal, Naman Goyal, Vishrav Chaudhary, Guillaume Wenzek, Francisco Guzm{\'{a}}n, Edouard Grave, Myle Ott, Luke Zettlemoyer, and Veselin Stoyanov.
\newblock Unsupervised cross-lingual representation learning at scale.
\newblock \emph{CoRR}, abs/1911.02116, 2019.
\newblock URL \url{http://arxiv.org/abs/1911.02116}.

\bibitem[Liu et~al.(2019)Liu, Ott, Goyal, Du, Joshi, Chen, Levy, Lewis, Zettlemoyer, and Stoyanov]{roberta}
Yinhan Liu, Myle Ott, Naman Goyal, Jingfei Du, Mandar Joshi, Danqi Chen, Omer Levy, Mike Lewis, Luke Zettlemoyer, and Veselin Stoyanov.
\newblock Roberta: {A} robustly optimized {BERT} pretraining approach.
\newblock \emph{CoRR}, abs/1907.11692, 2019.
\newblock URL \url{http://arxiv.org/abs/1907.11692}.

\bibitem[Ouyang et~al.(2021)Ouyang, Wang, Pang, Sun, Tian, Wu, and Wang]{erniem}
Xuan Ouyang, Shuohuan Wang, Chao Pang, Yu~Sun, Hao Tian, Hua Wu, and Haifeng Wang.
\newblock {ERNIE}-{M}: Enhanced multilingual representation by aligning cross-lingual semantics with monolingual corpora.
\newblock In \emph{Proceedings of the 2021 Conference on Empirical Methods in Natural Language Processing}, pages 27--38, Online and Punta Cana, Dominican Republic, November 2021. Association for Computational Linguistics.
\newblock \doi{10.18653/v1/2021.emnlp-main.3}.
\newblock URL \url{https://aclanthology.org/2021.emnlp-main.3}.

\bibitem[Lenz et~al.(2021)Lenz, Tashu, and Horv{\'a}th]{Marc2021}
Marc Lenz, Tsegaye~Misikir Tashu, and Tom{\'a}{\v{s}} Horv{\'a}th.
\newblock Learning inter-lingual document representations via concept compression.
\newblock In Hujun Yin, David Camacho, Peter Tino, Richard Allmendinger, Antonio~J. Tall{\'o}n-Ballesteros, Ke~Tang, Sung-Bae Cho, Paulo Novais, and Susana Nascimento, editors, \emph{Intelligent Data Engineering and Automated Learning -- IDEAL 2021}, pages 268--276, Cham, 2021. Springer International Publishing.
\newblock ISBN 978-3-030-91608-4.

\bibitem[Lenz(2021)]{Map}
Marc Lenz.
\newblock Learning multilingual document representations, 2021.

\bibitem[Steinberger et~al.(2006)Steinberger, Pouliquen, Widiger, Ignat, Erjavec, Tufis, and Varga]{JRC-Acquis}
Ralf Steinberger, Bruno Pouliquen, Anna Widiger, Camelia Ignat, Tomaz Erjavec, Dan Tufis, and D{\'{a}}niel Varga.
\newblock The jrc-acquis: {A} multilingual aligned parallel corpus with 20+ languages.
\newblock \emph{CoRR}, abs/cs/0609058, 2006.
\newblock URL \url{http://arxiv.org/abs/cs/0609058}.

\end{thebibliography}





% \end{thebibliography}

\end{document}